\documentclass[3p,twocolumn]{elsarticle}

\journal{Neurocomputing}

\usepackage{amsmath,amsfonts}
\usepackage{url}
\usepackage{graphicx}
\graphicspath{{figures/}}
\usepackage{amssymb}
\usepackage{bm}
\usepackage{booktabs}
\usepackage{siunitx}
\renewcommand{\vec}[1]{\bm{\mathrm #1}}

\begin{document}

\begin{frontmatter}

\title{Offset equivariant networks and their applications}

\author[1]{Marco Cotogni}
\ead{marco.cotogni01@universitadipavia.it}
\author[1]{Claudio Cusano\corref{cor1}}
\ead{claudio.cusano@unipv.it}
\cortext[cor1]{Corresponding author}

\address[1]{Dep.\ of Electrical, Computer and Biomedical Engineering, University of Pavia, Via Ferrata 1, Pavia, 27100, Italy}

\begin{abstract}
In this paper we present a framework for the design and implementation of offset equivariant networks, that is, neural networks that preserve in their output uniform increments in the input.  In a suitable color space this kind of networks achieves equivariance with respect to the photometric transformations that characterize changes in the lighting conditions.  We verified the framework on three different problems: image recognition, illuminant estimation, and image inpainting. Our experiments show that the performance of offset equivariant networks are comparable to those in the state of the art on regular data.  Differently from conventional networks, however, equivariant networks do behave consistently well when the color of the illuminant changes.
\end{abstract}

\end{frontmatter}


\section{Introduction}
\label{sec:intro}

Many recent studies focused on the role of equivariance in neural networks~\cite{ravanbakhsh2017equivariance,kondor2018generalization,yarotsky2021universal}. Briefly, a network (or one of its components) is equivariant with respect to a group of transformations when the application of one of such transformations to the input determines a predictable change in the output. For instance, convolutions are equivariant with respect to translations in the plane, since a translation in the input corresponds to an equal translation in the output.

Equivariance is a very important property, since it allows to impose an inductive bias~\cite{mitchell1980need}.  For instance, in convolutional networks it allows to exploit the knowledge that objects and patterns may appear in different locations of an image~\cite{lecun1998gradient,krizhevsky2012imagenet}.  In graph neural networks it makes the model independent on the order used to label the vertices in the graphs~\cite{maron2018invariant}.  In 3D processing equivariance allows to generalize convolutions to 3D meshes~\cite{de2020gauge}.

In computer vision various models of convolutional networks have been proposed to achieve equivariance with respect to geometric transformations such as translations, rotations, scalings and reflections~\cite{cohen2016group,cohen2016steerable}, and this made it possible to obtain impressive results in a variety of applications~\cite{de2020stability,de2021adapting,chiang2019wavelet,de2022convergent,mujica2021superpixels,lopez2020multi,yan2020deep,yan2020depth,yan2021task,yan2021precise,yan2022age}.
As far as we know, the problem of achieving equivariance with respect to \emph{photometric} transformations, such as those caused by variations in the lighting conditions or in the acquisition device, has yet to be explored.  An image recognition network equivariant with respect to photometric transformations, for instance, would make it possible to obtain accurate predictions even when the acquisition device is not the same used during training, or when facing a new unforeseen kind of illumination. These kinds of variations are often dealt with the application of normalization techniques (e.g. color balancing) at the cost of discarding potentially useful information~\cite{bianco2017improving}.

\begin{figure}
    \centering
    \includegraphics{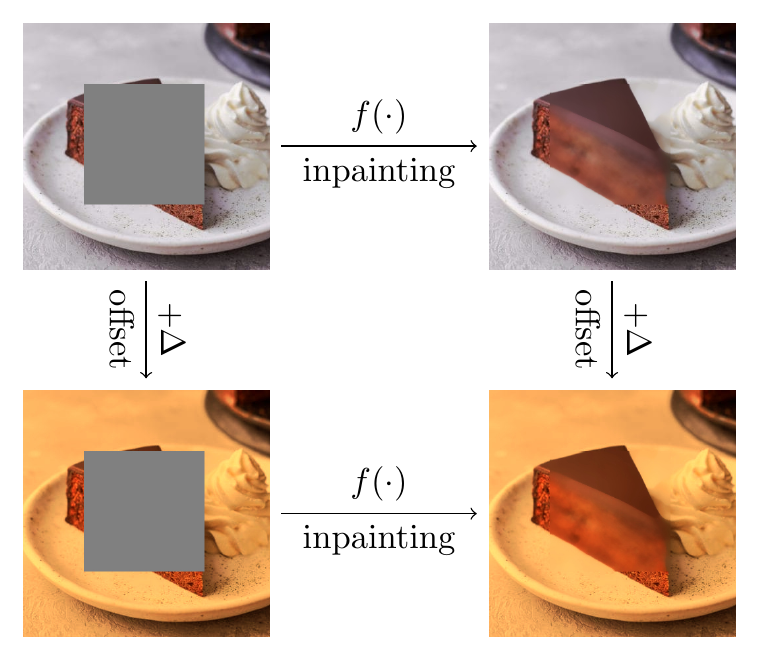}
    \caption{Representation of offset equivariance in the case of inpainting of color images.  Under the hypothesis that $f$ is an offset equivariant neural network implementing image inpainting, a change of illuminant and the application of $f$ can occur in any order without affecting the result.}
    \label{fig:inpainting-example}
\end{figure}

In this paper we propose a framework that allows to design neural networks that are equivariant to a specific group of photometric transformations.
In particular, we consider transformations consisting in the addition of a uniform offset to the input. 
For grayscale images the offset corresponds to a variation in the brightness.
For color images, provided that a suitable color space is used, the offset corresponds to a variation in the color of the illuminant of the scene.
Figure~\ref{fig:inpainting-example} shows the desired effect in the case of a neural network trained for image inpainting: a change in the color of the illuminant in the input image makes the network produce an output identical to the original output modified by the same change in the illuminant.

Many neural architectures can be modified according to the framework to achieve offset equivariance.  Here we explore three applications:
\begin{enumerate}[(i)]
    \item in image recognition, to achieve invariance with respect to global lighting conditions;    
    \item in illuminant estimation~\cite{gijsenij2011computational}, to make it robust to rare and unseen illuminants;
    \item in automatic image editing (inpainting~\cite{guillemot2013image}, compositing~\cite{tsai2017deep}, etc.) to make the methods automatically adjust to the illumination in the input image.  
\end{enumerate}

The rest of the paper is organized as follows: Section~\ref{sec:equivariance} defines offset equivariance and its properties; it also presents a strategy to convert conventional neural networks into equivariant ones.  Section~\ref{sec:ill_equiv} explains how offset equivariance can be used to achieve equivariance to changes in illumination. Section~\ref{sec:applications} describes three different experiments in which we explore the three kinds of applications listed above.  Finally, Section~\ref{sec:conclusions} discusses the results obtained and outlines some possible directions for future research.

\section{Offset equivariant networks}
\label{sec:equivariance}

In this section we propose a definition for \emph{offset equivariant} functions, show some of their useful properties, and use them to define the layers that will compose offset equivariant neural networks.

Intuitively, an \emph{offset equivariant} network has the property of preserving the global ``level'' of the input signal.  If we uniformly increase each component of the input $\vec{x}$ by the value $\Delta$ then all the components of the output will increase by the same amount.  More formally, the property is summarized by the following equation, which must hold true for all $\vec{x} \in \mathbb{R}^m$  and for all $\Delta \in \mathbb{R}$:
\begin{equation}
    \label{eq:equivgray}
    f(\vec{x} + \vec{1}_m \Delta) = f(\vec{x}) + \vec{1}_n \Delta, 
\end{equation}
where $f:\mathbb{R}^m \to \mathbb{R}^n$ is a function representing a neural network (or one of the layers composing it), and $\vec{1}_m, \vec{1}_n$ denote vectors of $m$ and $n$ unitary components, respectively.  For instance, when $\vec{x}$ is a gray-level image, adjusting the brightness of all pixels by $\Delta$ causes an equal increase in the output.

For color images enforcing Equation~(\ref{eq:equivgray}) is not really useful because photometric transformations may unevenly affect the color channels.  To deal with color images we assign input and output features to groups. Typically we will have a separate group for each color channel.  We then define offset equivariance as the property of preserving uniform variations of features in the same group.  This can be expressed as follows:

\begin{equation}
    \label{eq:equiv}
    f(\vec x + G_m \vec \Delta) = f(\vec x) + G_n \vec \Delta,
\end{equation}
where $\vec\Delta \in \mathbb{R}^g$ is a vector of offsets, one for each group, and where $G_m$ is a $m \times g$ matrix representing the assignment of the $m$ input features to the $g$ groups:
\begin{equation}
    (G_m)_{ij} = \begin{cases}
    1 & \text{if the \(i\)-th feature is in the \(j\)-th group}, \\
    0 & \text{otherwise}.
    \end{cases}
\end{equation}
Similarly, the $n \times g$ matrix $G_n$ encodes the assignment of the $n$ output features to the $g$ groups.  Note that the definition of equivariance depends on the assignment matrices.  To simplify the notation, in the rest of the paper we will assume that compatible assignments are used when multiple equivariant functions are involved.

An important property of offset equivariant functions is that they are closed under functional composition.  If $f_1:\mathbb{R}^m \to \mathbb{R}^n$ and $f_2:\mathbb{R}^p \to \mathbb{R}^m$ are both offset equivariant then their composition $h = f_1 \circ f_2$ is also offset equivariant.  This follows immediately from Equation (\ref{eq:equiv}):
\begin{equation}
    \begin{split}
    h(\vec x + G_p \vec\Delta) 
    &= f_1(f_2(\vec x + G_p \vec\Delta))
    = f_1(f_2(\vec x) + G_m \vec\Delta) \\
    &= f_1(f_2(\vec x)) + G_n \vec\Delta
    = h(\vec x) + G_n \vec\Delta.
    \end{split}
\end{equation}

Even though linear combinations in general do not preserve offset equivariance, affine linear combinations do exactly that.  Given the equivariant functions $f_1$, $f_2$, \dots, $f_k$ and the coefficients $\alpha_1$, $\alpha_2$, \dots, $\alpha_k$, the function
$h$ defined as:
\begin{equation}
    \label{eq:combination}
    h(\vec x) = \sum_{i=1}^k \alpha_i f_i(\vec x),
\end{equation}
is offset equivariant if and only if $\sum_{i=1}^k \alpha_i = 1$.  In fact we have:
\begin{equation}
    \begin{split}
    h(\vec x + G_m \vec\Delta)
    &= \sum_{i=1}^k \alpha_i f_i(\vec x + G_m\vec\Delta) \\
    & = \sum_{i=1}^k \alpha_i\left( f_i(\vec x) + G_n\vec\Delta \right) \\
    &= \left(\sum_{i=1}^k \alpha_i f_i(\vec x)\right) + \left(\sum_{i=1}^k \alpha_i \right) G_n\vec\Delta \\ 
    & = h(\vec x) + G_n \vec \Delta.
    \end{split}
\end{equation}

These properties suggest a possible strategy for the design of offset equivariant networks that consists in defining them as the composition and combination of the offset equivariant layers that will be defined in the following sections.

\subsection{Linear layers}

Most neural architectures include one or more linear layers.  They can be found in multi-layer perceptrons~\cite{rosenblatt1958perceptron}, in convolutional neural networks~\cite{krizhevsky2012imagenet}, in graph neural networks~\cite{scarselli2008graph}, in attention mechanisms~\cite{vaswani2017attention}, and many more.
Linear layers are defined by means of a $n \times m$ weight matrix $W$ and a $n$-dimensional bias vector $\vec b$:
\begin{equation}
    f(\vec x) = W \vec x + \vec b.
\end{equation}
The introduction of an offset leads to:
\begin{equation}
    f(\vec x + G_m \vec\Delta)
    = W \vec x + W G_m \vec\Delta + \vec b
    = f(\vec x) + W G_m \vec\Delta,
\end{equation}
which, combined with Equation~(\ref{eq:equiv}), gives us 
\begin{equation}
W G_m \vec\Delta = G_n \vec\Delta.
\end{equation}
Since the above must hold true independently on $\vec\Delta$, we can summarize offset equivariance for linear layers with the linear constraint:
\begin{equation}
\label{eq:linear-constraint}
W G_m = G_n,
\end{equation}
which is feasible if and only if  all the groups to which $G_n$ assigns at least one feature also have at least one feature assigned to by $G_m$.  Note that the equivariance of $f$ does not depend on the bias vector $\vec b$, which can remain a free parameter.
In practice, $W$ could be reparametrized to satisfy the constraint (\ref{eq:linear-constraint}).  An alternative is to work with a standard linear layer, and project $W$ onto the constraint after each iteration of backpropagation.  The projected weights $\hat{W}$ can be computed as follows:
\begin{equation}
\label{eq:weight-projection}
\hat{W} = W - (W G_m - G_n) G_m^+,
\end{equation}
where $G_m^+$ is the pseudoinverse of $G_m$  (due to the regular structure of $G_m$ its pseudoinverse has the simple form $G_m^+ = N^{-1} G_m^T$, where $N$ is the diagonal matrix having as elements the number of features in each group).  The weights of the linear layer are updated by alternating the usual backpropagation steps and the projection defined in (\ref{eq:weight-projection}).

\subsection{Convolutional layers}

Since convolutions are linear operators they can be treated in the same way discussed in the previous section.  Here we consider in particular the two-dimensional case, but the generalization to other numbers of dimensions is straightforward.  Given a set of coefficients $\mathcal{W} \in \mathbb{R}^{k \times k \times n \times m}$, organized as a $k \times k$ array of linear operators $\mathcal{W}_{ij} \in \mathbb{R}^{n \times m}$, the convolution for the input image $\vec x$ is denoted as:
\begin{equation}
f(\vec x) = \mathcal{W} \star \vec x.
\end{equation}
The pixel value at location $(i,j)$ is the $m$-dimensional vector $\vec x_{ij}$ and the $n$-dimensional output value at location $(i,j)$ is defined as:
\begin{equation}
f(\vec x)_{ij} = \sum_{s=1}^{k} \sum_{t=1}^{k} \mathcal{W}_{st}\vec{x}_{i + s - 1, j + t - 1}.
\end{equation}

Let $\vec x'$ be the image obtained by adding the offset $G_m \vec\Delta$ to each pixel of $\vec x$, then the corresponding output value is:
\begin{equation}
\begin{split}
f(\vec x')_{ij}
&= \sum_{s=1}^{k} \sum_{t=1}^{k} \mathcal{W}_{st}\vec{x}'_{i + s - 1, j + t - 1}\\
&= \sum_{s=1}^{k} \sum_{t=1}^{k} \mathcal{W}_{st}(\vec{x}_{i + s - 1, j + t - 1} + G_m \vec\Delta)\\
&= \sum_{s=1}^{k} \sum_{t=1}^{k} \mathcal{W}_{st}\vec{x}_{i + s - 1, j + t - 1} + \left(\sum_{s=1}^{k} \sum_{t=1}^{k} \mathcal{W}_{st}\right) G_m \vec\Delta\\
&= f(\vec{x})_{ij} + \left(\sum_{s=1}^{k} \sum_{t=1}^{k} \mathcal{W}_{st}\right) G_m \vec\Delta.
\end{split}
\end{equation}
Therefore, in order to achieve offset equivariance we need to have: 
\begin{equation}
\left(\sum_{s=1}^{k} \sum_{t=1}^{k} \mathcal{W}_{st}\right) G_m \vec\Delta = G_n \vec\Delta,
\end{equation}
for all $\vec\Delta \in \mathbb{R}^g$, which corresponds to the following linear constraint:
\begin{equation}
\label{eq:conv_constraint}
\left(\sum_{s=1}^{k} \sum_{t=1}^{k} \mathcal{W}_{st}\right) G_m = G_n.
\end{equation}
A projection $\hat{\mathcal{W}}$ satisfying (\ref{eq:conv_constraint}) is the following:
\begin{equation}
\label{eq:conv-projection}
\hat{\mathcal{W}}_{ij} = \frac{1}{k^2} \left(\mathcal{W}_{ij} - (\mathcal{W}_{ij} G_m - G_n) G_m^+ \right).
\end{equation}
and the weights of the convolutional layer are updated by alternating the usual backpropagation steps and the projection defined in (\ref{eq:conv-projection}).

In order to preserve offset equivariance, When padding is used the padding strategy must be of the kind that uses the input data, for instance by replicating the values on the border.  Padding with zeros, or with other constant values would prevent equivariance.  Strided and dilated convolutions require no special treatment.

\subsection{Pooling layers}
Standard average and max pooling are already offset equivariant operators, under the same constraint on padding imposed by convolutions.

\subsection{Group pooling and non-linear layers}
Activation functions, and other non-linear layers such as dropout~\cite{srivastava2014dropout} and batch normalization \cite{ioffe2015batch} are not offset equivariant.  However, we can turn them equivariant by using auxiliary \emph{group pooling functions}.  A group pooling function is an equivariant function $\varphi:\mathbb{R}^m \to \mathbb{R}^g$ aggregating feature values by group.  With respect to the definition (\ref{eq:equiv}) we add the requirement
that $G_n$ is the $g \times g$ the identity matrix. This way, for group pooling functions we have:
\begin{equation}
    \label{eq:grouppool}
    \varphi(\vec x + G_m \vec\Delta) = \varphi(\vec x) + \vec\Delta.
\end{equation}
Are examples of group pooling functions the average over the groups, the maximum, and the minimum.  Another group pooling function is the linear combination of the features, provided that the coefficients satisfy the constraint (\ref{eq:linear-constraint}).

Group pooling functions can be used to make offset equivariant functions from any other function.  Let $f$ be any $\mathbb{R}^m \to \mathbb{R}^n$ function, and let $\varphi_1$, $\varphi_2$ be two group pooling functions, then the function $h$, defined as:
\begin{equation}
    \label{eq:makeequiv}
    h(\vec x) = f(\vec x - G_m \varphi_1(\vec x)) + G_n \varphi_2(\vec x),
\end{equation}
is offset equivariant.  The demonstration follows immediately from (\ref{eq:equiv}) and (\ref{eq:grouppool}):
\begin{equation}
\begin{split}
    h(\vec x + G_m \vec\Delta)
    &= f(\vec x + G_m \vec\Delta - G_m \varphi_1(\vec x + G_m \vec\Delta)) \\ &+ G_n \varphi_2(\vec x + G_m \vec\Delta) \\
    &= f(\vec x + G_m \vec\Delta - G_m \varphi_1(\vec x ) -  G_m \vec\Delta) \\ &+ G_n \varphi_2(\vec x) + G_n \vec\Delta\\
    &= f(\vec x - G_m \varphi_1(\vec x)) + G_n \varphi_2(\vec x)  + G_n \vec\Delta \\
    &= h(\vec x) + G_n \vec\Delta.
\end{split}
\end{equation}

As an example, let's apply this technique to the ReLU activation function, one of the most widely used components of deep neural networks.  Let $\varphi$ be a group pooling function, then the application to ReLUs of (\ref{eq:makeequiv}) with $\varphi_1 = \varphi_2 = \varphi$ leads to:
\begin{equation}
    \begin{split}
    h(\vec x) 
    &= \operatorname{relu}(\vec{x} - G_m \varphi(\vec x)) + G_m\varphi(\vec x) \\
    &= \max(\vec{x} - G_m \varphi(\vec x), \vec 0) + G_m\varphi(\vec x)  \\
    &= \max(\vec{x} , G_m\varphi(\vec x)).
    \end{split}
\end{equation}
In particular, when $\varphi$ is a linear function this resembles a maxout network~\cite{goodfellow2013maxout} (in fact the maxout activation function is naturally offset equivariant).


A working implementation of the main offset equivariant layers is available online at the address \url{https://github.com/claudio-unipv/offset-equivariant}.

\section{Illuminant equivariance}
\label{sec:ill_equiv}
The addition of a constant offset is not the typical transformation that is applied to color images.  Multiplication is more useful since, according to the the von Kries diagonal model~\cite{von1902chromatic}, it corresponds to the chromatic variation caused by a global change in the illuminant:
\begin{equation}
\label{eq:vonkries}
\begin{pmatrix}y_r\\y_g\\y_b\end{pmatrix} =
\begin{bmatrix}
I_r & 0 & 0 \\
0 & I_g & 0 \\
0 & 0 & I_b
\end{bmatrix} \cdot
\begin{pmatrix}x_r\\x_g\\x_b\end{pmatrix},
\end{equation}
where $x_r, x_g, x_b$ is the color of a pixel in the linear RGB space under a canonical neutral illuminant, and where $y_r, y_g, y_b$ is the color of the same pixel under the illuminant $I_r, I_g, I_b$.

A logarithmic transformation
\begin{equation}
\label{eq:logrgb}
x'_c = \log \frac{1}{x_c}, \; \;
y'_c = \log \frac{1}{y_c}, \; \;
I'_c = \log \frac{1}{I_c}, 
\; \; c \in \{r, g, b\},
\end{equation}
makes the model additive:
\begin{equation}
\begin{pmatrix}y'_r\\y'_g\\y'_b\end{pmatrix} =
\begin{pmatrix}I'_r\\I'_g\\I'_b\end{pmatrix} +
\begin{pmatrix}x'_r\\x'_g\\x'_b\end{pmatrix}.
\end{equation}
In this logarithmic RGB color space offset equivariance corresponds to the equivariance with respect to the transformation caused by a change in the illuminant.

In performing the transformation, there is one detail we need to pay attention to:
if the input component is exactly zero, then the transformed one would be $+\infty$.
To prevent it, all values in the original linear space are clipped from below to a small positive value $\epsilon$ (in all the experiments we set $\epsilon = 2 \times 10^{-4}$). 

\section{Applications}
\label{sec:applications}

To verify the feasibility of offset equivariant networks we selected three applications in which convolutional neural networks represent the state-of-the-art solution.  The three applications are: image recognition, illuminant estimation and image inpainting.  For each one we compared offset equivariant networks with the standard counterparts taken from the state of the art.
More precisely, for each application we took one or more CNN architectures from the literature and we modified them to obtain their equivariant versions.  We then trained the original and the modified versions on reference datasets.  The performance of the variants are compared on the reference test sets with and without the application of a data augmentation procedure corresponding to a random change of the global illuminant.  Note that such an augmentation is never used during training.

To make the networks equivariant we replaced the original components with those 
described in Section~\ref{sec:equivariance}.  We used three feature groups (associated to the red, green and blue color channels) and assignment matrices that place an equal number of features in each group:
\begin{equation}
    (G_m)_{ij} = \begin{cases}
        1 & \text{if} \; \left\lceil \frac{3i}{m} \right\rceil = j, \\
        0 & \text{otherwise}.
    \end{cases}
\end{equation}
This requires that the number of features is a multiple of three.
Where needed, we modified the original architectures by adjusting the number of features at each layer to the closest multiple of three.  

For convolutions and linear layers we just enforced the constraint (\ref{eq:linear-constraint}) by projecting the weights onto it.  Pooling layers were kept unchanged. ReLUs and batch normalization layers were adapted by applying Equation (\ref{eq:makeequiv}) with average group pooling.

The following sections provide additional details and report the results obtained for each task.

\subsection{Image recognition}
\label{sec:image-recognition}

Image classification is probably the most thoroughly explored application of convolutional neural networks.  Many important advancements in deep learning have been proposed and assessed while addressing this problem.  

A recurrent goal in image classification is to achieve robustness to variations in the input.  Data augmentation~\cite{perez2017effectiveness} is a widely used technique to achieve it.  Here we will verify if offset equivariant convolutional networks are robust with respect to large variations in the global color distribution even when trained without any specific data augmentation.

To do so, we considered the ResNet~\cite{he2016deep} family of convolutional networks, which is considered one of the most performing options for image classification, and one of the most widely studied.  ResNets include convolutions, ReLU activation functions, batch normalization layers and a final average pooling followed by a fully connected layer.  All these building blocks can be made offset equivariant by the procedures described in the previous sections.  The only conversion that is not straightforward is that of the residual blocks after which the architecture is named:
\begin{equation}
    \label{eq:residual}
    \vec y = f(\vec x) + \vec x.
\end{equation}
Residual blocks directly combine the function $f$ (that can be made equivariant) and the identity function (that is trivially equivariant).
However, the sum of two equivariant functions is not equivariant!
In fact, Equation (\ref{eq:combination}) requires that the coefficients in the linear combination sum up to one.  To preserve offset equivariance we defined the following modified residual block:
\begin{equation}
    \label{eq:mod-residual}
    \vec y = f(\vec x) + \vec x - G_m \varphi(\vec x),
\end{equation}
and we used the average group pooling as function $\varphi$.

To assign the features uniformly to the three groups, the number of channels computed by each convolution is adjusted to the closest multiple of three.  
The number of output scores, computed by the final fully connected layer is instead set to triple the number of classes, so that one score is computed for each class and for each  group of features.
The result of all the modifications is an offset equivariant version of the original ResNet.

Offset equivariance can be used to achieve illuminant invariance under the von Kries model (\ref{eq:vonkries}).  The input image is assumed to be in the sRGB color space.  The sRGB gamma is removed, and the image is converted in the log RGB space by applying Equation (\ref{eq:logrgb}).  The converted image is normalized by using the log RGB mean and standard deviations computed on the training set.  Then the CNN is applied obtaining a set of $k \times 3$ scores (where $k$ is the number of classes).  The scores are reduced to a $k$-dimensional vector by averaging over the groups and, finally, the softmax operator is applied to obtain the posterior probabilities associated to the $k$ classes.

More precisely, let $Z \in \mathbb{R}^{k \times 3}$ be the output of  the equivariant CNN for an image $\vec x$, where $(z_{i1}, z_{i2}, z_{i3})$ is the triplet of scores for class $i$.  A change of illuminant in the log RGB space corresponds to the addition of the same offset $\vec I$ to all the pixels.
Since the network is offset equivariant, the scores for the modified image will be $(z_{i1} + I_1, z_{i2} + I_2, z_{i3} + I_3)$.  The posterior probability $p_i$ for class $i$ is obtained by applying the softmax operator to the average of the scores:
\begin{equation}
    \label{eq:sotfmax-illuminant}
    \begin{split}
    p_i
    &= \frac{e^{\frac{1}{3}\left( z_{i1} + I_1 + z_{i2} + I_2 + z_{i3} + I_3 \right)}}
    {\sum_{j=1}^k e^{\frac{1}{3}\left( z_{j1} + I_1 + z_{j2} + I_2 + z_{j3} + I_3 \right)}} \\
    &= \frac{e^{\frac{1}{3} (I_1 + I_2 + I_3)} e^{\frac{1}{3}\left( z_{i1} + z_{i2} + z_{i3} \right)}}
    {e^{\frac{1}{3} (I_1 + I_2 + I_3)} \sum_{j=1}^k e^{\frac{1}{3}\left( z_{j1} + z_{j2} + z_{j3} \right)}} \\
    &= \frac{e^{\frac{1}{3}\left( z_{i1} + z_{i2} + z_{i3} \right)}}
    {\sum_{j=1}^k e^{\frac{1}{3}\left( z_{j1} + z_{j2} + z_{j3} \right)}},
    \end{split}
\end{equation}
which does not depend on $\vec I$.  The final probability estimates can be used to predict the most probable class label, or during training as part of the cross entropy classification loss.

\subsubsection{CIFAR-10}
\label{sec:cifar}
The first dataset we considered is the Canadian Institute for Advanced Research-10 dataset (CIFAR-10)~\cite{krizhevsky2009learning}. The CIFAR-10 dataset is a subset of a larger dataset of tiny images. Each of these images have a dimension of $32 \times 32$ pixels. This labeled subset of images is composed of 10 classes of objects: airplane, automobile, bird, cat, deer, dog, frog, horse, ship, and truck. The classes do not overlap (i.e. each image depicts an object that belongs only to a single class). The dataset includes \num{6000} images for each of the ten classes, for a total of \num{60000} images. The dataset has been split in training (50k) and test set (10k) by the authors.

We replicated the experiment on CIFAR-10 described in the paper presenting the ResNet family of networks.  In particular we considered the network composed of 20 layers (i.e. nine residual blocks).  We trained the original model by replicating the setup described in the paper obtaining a level of accuracy matching the reported one (8.67\% of classification error on the test set).

The experiment was repeated by using an offset equivariant version of the network, obtaining a very similar test error (8.83\%).

In order to verify the capability of the offset equivariant network in tolerating large variations in the illuminant, we repeated the evaluation step by distorting the input image as follows:
\begin{enumerate}
\item the sRGB gamma is removed;
\item a random color of the illuminant is generated in the HSV color space by randomly taking the hue $H$;  $V$ is set to the maximum value, and the saturation $S$ is left as a tunable parameter to make it possible to adjust the amount of chromatic distortion;
\item the illuminant color is converted to the RGB space and applied to the whole image (Equation (\ref{eq:vonkries}));
\item the sRGB gamma is restored.
\end{enumerate}

\begin{figure}
    \centering
    \includegraphics[width=\linewidth]{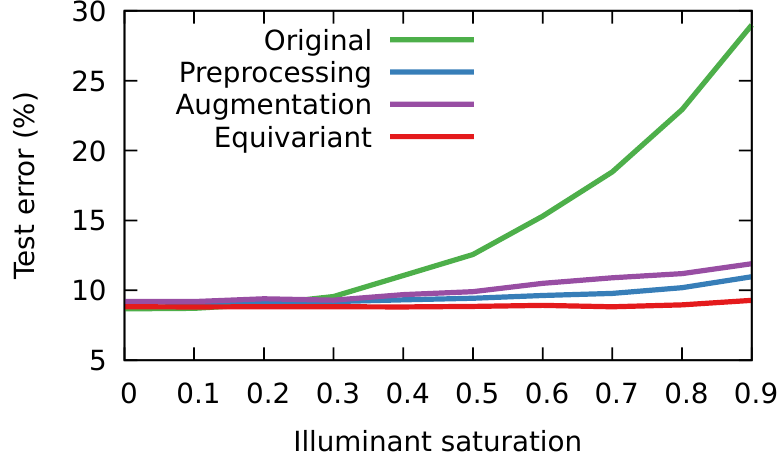}
    \caption{Error rates of original and equivariant ResNet-20 on CIFAR-10 test images under an illuminant with random hue and variable saturation.
    For the original network, the results obtained with preprocessing and data augmentation are also reported.}
    \label{fig:cifar}
\end{figure}

Figure~\ref{fig:cifar} reports the test accuracies obtained by the original and the equivariant model, as a function of the saturation of the generated illuminant.  
The comparison includes the performance obtained with two alternative strategies to make the original network robust to changes in illumination:
\begin{itemize}
\item data augmentation, in which color jittering is applied during training.  We tried different amounts of random brightness, contrast, hue and saturation and kept the combination with the best average performance on the test set.
\item Preprocessing, in which an automatic color balancing algorithm is applied to training and test images.  Among the many methods in the state of the art we used Quasi-unsupervised  color constancy~\cite{bianco2019quasi}, since for training it does not require information about the actual color of the illuminant in the scene, and since it can work in the standard sRGB space (most methods work in the device RAW space).  We used the version available online pretrained on the ImageNET data in the unsupervised scenario.
\end{itemize}

The plot shows how, for small saturation values, the performance of different networks is very similar.
As soon as the saturation goes above 0.3 the test error of the original ResNet starts to increase and it monotonically diverges from the error rate obtained under the neutral ($S = 0$) illuminant.  The behavior of the offset equivariant version is, instead, very stable.  The same error rate ($\pm 0.2\%$) is obtained for values of saturation up to 0.8.  Only At $S = 0.9$ there is a noticeable decrease in the accuracy (9.38\% of error) and only at $S = 1.0$ the performance are seriously degraded due to the complete elimination of the information from one or more color channels (28.17\% of test error, not reported in the plot to make it easier to read).  Both preprocessing and data augmentation demonstrated to be valid strategies, but the offset equivariant version is even more accurate, does not require any time consuming tuning of the parameters (as data augmentation did) and does not require the training of a complicated additional model for preprocessing (the one we used includes more than 54 millions of learnable parameters).

The results of the test are summarized in Table~\ref{tab:cifar}.  The small difference in the number of parameters is due to the adjustment of the number of channels to make them multiples of three.
\begin{table}
    \centering    
    \begin{tabular}{lcccc}
        \toprule
        & & \multicolumn{3}{c}{Test error  (\%)} \\
        Strategy & \# par. & $S~0.0$ & $S~0.5$ & $S~0.9$ \\
        \midrule
        Original & \num{270410} & \textbf{8.67} & 12.6 & 29.0 \\
        Preprocessing & \num{270410} & 9.19 & 9.43 & 11.0 \\
        Augmentation & \num{270410} & 9.20 & 9.91 & 11.9 \\
        Equivariant & \num{267294} & 8.88 & \textbf{8.85} & \textbf{9.29} \\
        \bottomrule
    \end{tabular}
    \caption{Comparison of original and offset equivariant ResNet-20 on CIFAR-10 test images.  For the original ResNet, the versions trained with preprocessing and data augmentation are also compared.  For the version with preprocessing only the parameters in the ResNet have been counted. The preprocessing module includes about 54 millions of extra parameters.
    Three levels of distortions are considered: no distortion ($S=0.0$), moderate  ($S=0.5$) and strong distortion ($S=0.9$).}
    \label{tab:cifar}
\end{table}

\begin{figure}
    \centering
    \begin{tabular}{r@{\hspace{1pt}}l}
    & \begin{tabular}{*{5}{p{0.095\textwidth}@{}}}
         $S = 0$ & 0.2 & 0.4 & 0.6 & 0.8 \\
    \end{tabular} \\
    & \includegraphics[width=0.47\textwidth]{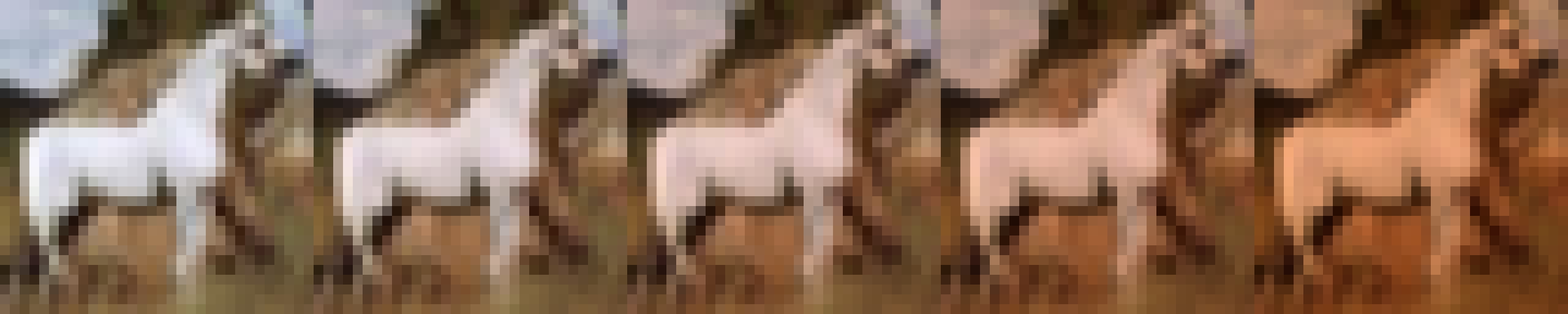} \\
    \rotatebox{90}{\scriptsize orig.}
    & \includegraphics[width=0.47\textwidth]{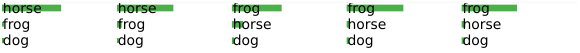} \\
    \rotatebox{90}{\scriptsize equiv.}
    & \includegraphics[width=0.47\textwidth]{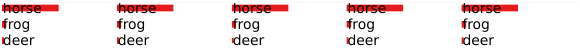} \\
    & \includegraphics[width=0.47\textwidth]{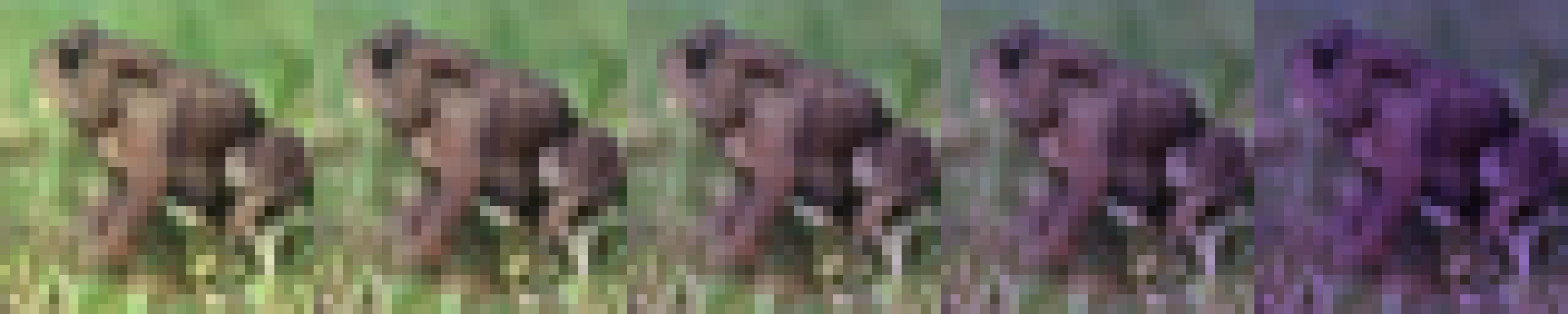} \\
    \rotatebox{90}{\scriptsize orig.}
    & \includegraphics[width=0.47\textwidth]{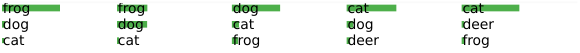} \\
    \rotatebox{90}{\scriptsize equiv.}
    & \includegraphics[width=0.47\textwidth]{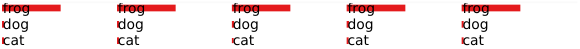} \\
    & \includegraphics[width=0.47\textwidth]{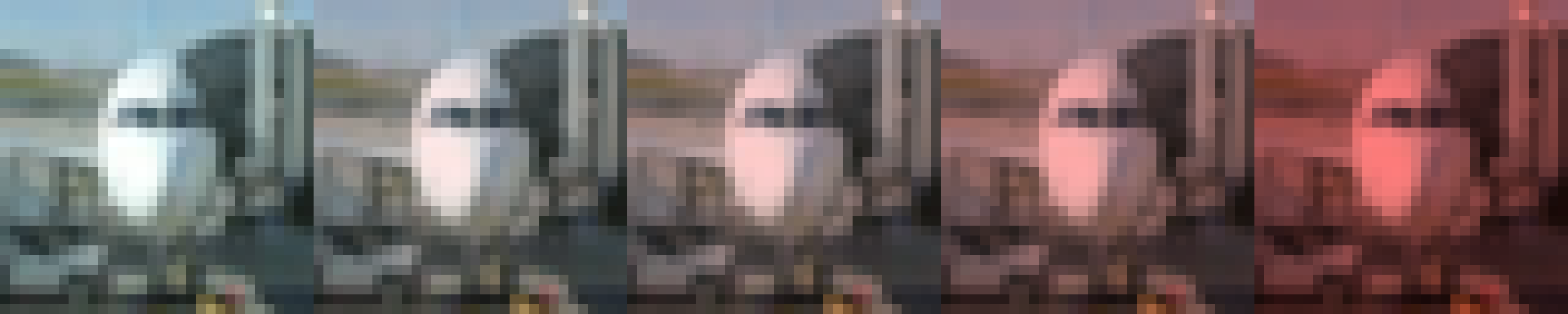} \\
    \rotatebox{90}{\scriptsize orig.}
    & \includegraphics[width=0.47\textwidth]{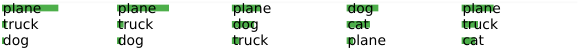} \\
    \rotatebox{90}{\scriptsize equiv.}
    & \includegraphics[width=0.47\textwidth]{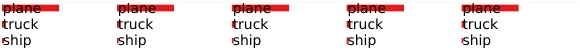} \\
    \end{tabular}
    \caption{Comparison of the predictions obtained on three images taken from the CIFAR-10 test set.  Each image has been distorted by the introduction of an artificial illuminant with saturation varying from 0.0 (no distortion) to 0.8.
    The top three classes predicted by the original and the equivariant ResNet-20 are reported below each image.  The lenghts of the bars is proportional to the posterior probabilities estimated by the networks.}
    \label{fig:cifar-examples}
\end{figure}

Figure \ref{fig:cifar-examples} shows how scores and predictions obtained with the original ResNet-20 are clearly affected by the introduction of an artificial illuminant, even when its saturation is mild.
On the other hand, the output of equivariant ResNet-20 is stable over the entire range of saturation values.
%

\subsubsection{ILSVRC 12}
\label{sec:ilsvrc}
For the second experiment on image recognition we used the dataset presented for the ImageNet Large Scale Visual Recognition Challenge 2012 (ILSVRC-12) \cite{ILSVRC15}. The dataset is composed of \num{1000} classes selected among all the available in the ImageNet dataset \cite{Deng2009} avoiding any overlap between them (in the original ImageNet dataset the classes are organized in a larger hierarchy where each class could be an ancestor with several children). Once the authors selected the \num{1000} classes they gathered the training images (about 1.2 millions) directly from the ImageNet dataset and the validation (50K) and testing (100k) images from several search engines such as Flickr and then filtered following the ImageNet guidelines.
\begin{table}
    \centering
    \begin{tabular}{lcccc}
        \toprule
        & & \multicolumn{3}{c}{Test error  (\%)} \\
        Strategy & \# par. & $S~0.0$ & $S~0.5$ & $S~0.9$ \\
        \midrule
        Original & \num{25.6}M & \textbf{24.59} & 26.09 & 31.17 \\
        Preprocessing & \num{25.6}M & 24.95 & 25.32 & 26.53 \\
        Augmentation & \num{25.6}M & 28.09 & 29.21 & 33.91 \\
        Equivariant & \num{28.8}M & 24.85 & \textbf{24.92} & \textbf{25.71} \\
        \bottomrule
    \end{tabular}
    \caption{Comparison of original and offset equivariant ResNet-50 on ILSVRC12 test images.  Results obtained by using preprocessing (without counting the extra parameters) and data augmentation on top of the original ResNet-50 are also reported.  
    Three levels of distortions are considered: no distortion ($S=0.0$), moderate  ($S=0.5$) and strong distortion ($S=0.9$).}
    \label{tab:resnet50_result}
\end{table}

On the basis of the experiments presented in the previous section, we trained an offset equivariant version of the ResNet-50. This version, has the same structure of the original Resnet-50~\cite{he2016deep} with the introduction of the equivariant layers presented before.
We followed the training procedure presented by He et al. for the ImageNet dataset (i.e. data augmentation with random resize crop, random horizontal flip, etc.). In order to train the model on a single Nvidia RTX 3080ti graphic card with 12 Gb of memory, we had to decrease the batch size to 48. We tested the model on the ILSVRC-12 test set and we reported the test error (24.85\%). We repeated the training procedure with the original version of the ResNet-50 (using the same batch size in order to have a fair comparison) obtaining the very similar test error of 24.59\% (see Table \ref{tab:resnet50_result} for the comparison of the results and number of parameters).

\begin{figure}
    \centering
    \includegraphics[width=\linewidth]{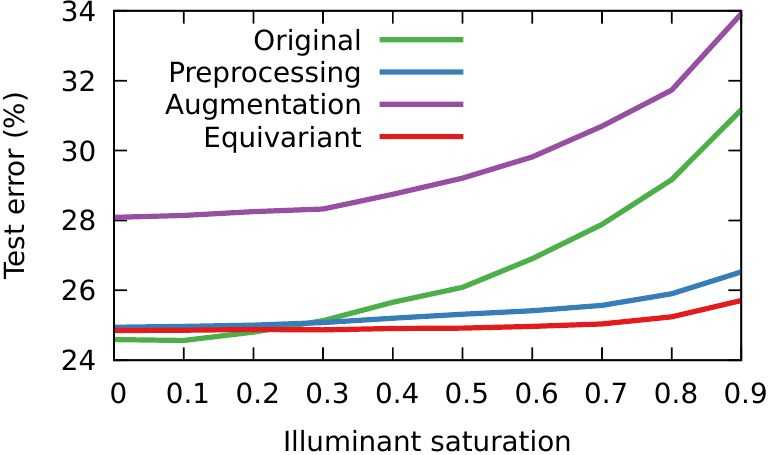}
    \caption{Error rates of original and equivariant ResNet-50 on ILSVRC12 test images under an illuminant with random hue and variable saturation.  For the original network, the results obtained with preprocessing and data augmentation are also reported.}
    \label{fig:ilsvrc}
\end{figure}

We then tested the robustness of the equivariant ResNet-50 and original ResNet-50 by varying the illuminant with the same procedure used in the experiment with CIFAR-10. Figure~\ref{fig:ilsvrc} reports the error rates as functions of the saturation. For small values of saturation the behavior of the two networks is very similar. As for the CIFAR-10 experiment, for values of saturation of 0.3, the original ResNet-50 error curve starts to grow. The equivariant ResNet-50 instead is stable up to values of saturation of 0.8 (test error 24.83\% $\pm 0.3$) and then slightly increase at $S=0.9$ (25.71\%).  We included in the comparison also the same preprocessing and data augmentation strategies aldready described for the experiments on the CIFAR-10 dataset.  With preprocessing we obtained quite good results (less than 1\% of additional error with respect to the equivariant network).  With data augmentation the results were not good.  This probably depends on the fact that we used the same augmentation parameters optimized for the CIFAR dataset since a complete tuning on ILSVRC would have been too computational intensive.

\subsection{Illuminant Estimation}
\label{sec:illuminant-estimation}

Illuminant estimation is the main step of most computational color constancy methods~\cite{gijsenij2011computational}.  It consists in the prediction of the color of the light source(s) in the scene displayed in  an image.  Once obtained, the estimated color of the illuminant can be discarded to reproduce the color constancy capability of the human visual system.  The result is a modified image in which colors appear as if the scene was taken under a canonical neutral illuminant.  For the correction step, most methods just reverse the transformation defined by the von Kries diagonal model (\ref{eq:vonkries}).

In the context of illuminant estimation, offset equivariance allows the prediction model to behave consistently with respect to changes in illuminantion.  
Let $\vec x$ and $\vec x'$ be two images representing the same scene taken under the illuminants $\vec I$ and $\vec I'$; in the logarithmic RGB space the von Kries model (\ref{eq:vonkries}) states that $x'_{ijc} = x_{ijc} - I_c + I'_c$. 
Under this hypothesis, the estimates $\hat{ \vec I}$ and $\hat{ \vec I}'$ of an offset equivariant model would preserve the difference between real and predicted illuminants ($\vec I - \vec I' = \hat{\vec I} - \hat{\vec I}'$) and this implies that the estimation error is independent on the actual color of the illuminant in the scene ($\vec I - \hat{\vec I} = \vec I' - \hat{\vec I}'$).

The use of Convolutional Neural Networks for illuminant estimation has been thoroughly explored~\cite{bianco2015color,lou2015color,oh2017approaching,bianco2019quasi}.  In this work we experimented with the Color Cerberus~\cite{savchik2019color}: a CNN that takes as input a $64 \times 64$ image and produces as output $k$ illuminant estimates (in the experiments we simply set $k=1$).  The network starts by augmenting each pixel with the global average.  The resulting six-channel image is subjected to a sequence of four modules consisting of $3 \times 3$ convolution and $2 \times 2$ max pooling.  Finally,
a $1 \times 1$ convolution, and two fully connected layers map the image to a three-dimensional illuminant estimate.  All convolutional and fully connected layers are followed by the ReLU activation function.

The conversion of the original Cerberus to its offset equivariant version is straightforward: it is enough to replace convolutions, ReLUs and fully connected linear layers with their offset equivariant versions.  The input image is converted to the log RGB space, and the output estimate is converted back in the linear RGB space.

\subsubsection{NUS dataset}
\label{sec:nus}
The dataset considered for this task is that proposed by researchers from the National University of Singapore (NUS)~\cite{cheng2014illuminant}. This dataset has been especially designed for the study of computational color constancy, and is composed of 1853 images taken with nine different commercial cameras.
The dataset contains indoor and outdoor scenes in which the authors inserted a color chart to make it possible to compute a reliable ground truth for the color of the illuminant (it is a common practice to cover the chart with a black patch during the experiments).

We trained the original and the equivariant versions of the color Cerberus by using the procedure described in the reference paper.  The two models have been evaluated by three-fold cross validation on the whole dataset using the reproduction angular error (\cite{finlayson2016reproduction}) as performance measure:
\begin{equation}
\label{eq:angerr}
E(\hat{\vec I}, \vec I) = \arccos \left( \frac{\hat I_r / I_r + \hat I_g / I_g + \hat I_b / I_b}
{\sqrt{3\left(\hat I^2_r / I^2_r + \hat I^2_g / I^2_g + \hat I^2_b / I^2_b \right)}} \right).
\end{equation}
Results are reported in Table~\ref{tab:nus}.
The accuracy of the two models is very similar, with the original version obtaining a slightly lower median error, but only for the original test images without distortions.
\begin{table}
    \centering    
    \begin{tabular}{lcccc}
        \toprule
        & & \multicolumn{3}{c}{Med. angular error (deg)} \\
        Network & \# par. & $S~0.0$ & $S~0.5$ & $S~0.9$ \\
        \midrule
        Original & \num{206345} & \textbf{1.93} & 11.5 & 30.3 \\
        Equivariant & \num{196668} & 2.03 & \textbf{2.30} & \textbf{3.44} \\
        \bottomrule
    \end{tabular}
    \caption{Comparison of original and offset equivariant color Cerberus on NUS images. Performance are measured in terms of median reproduction angular error and computed by three-fold cross validation on the whole dataset.  Three levels of distortions are considered: no distortion ($S=0.0$), moderate  ($S=0.5$) and strong distortion ($S=0.9$).
    }
    \label{tab:nus}
\end{table}

\begin{figure}
    \centering
    \includegraphics[width=\linewidth]{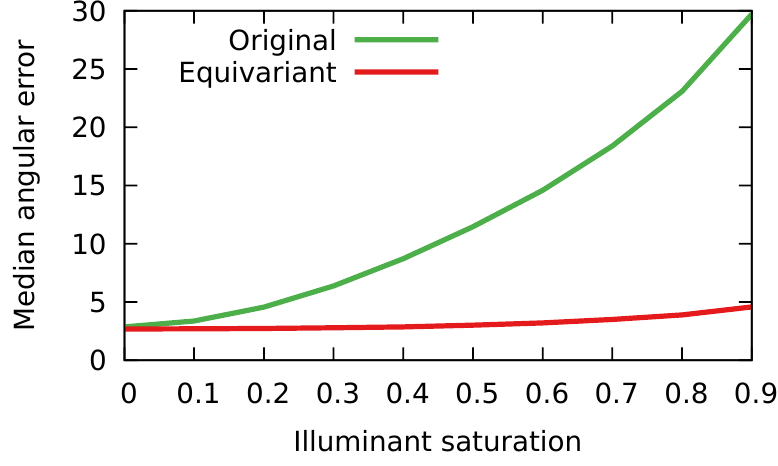}
    \caption{Median angular reproduction error on NUS images under an illuminant with random hue and variable saturation.}
    \label{fig:nus}
\end{figure}

\begin{figure}
    \centering
    \begin{tabular}{@{}r@{\hspace{1pt}}l@{}}
    & \begin{tabular}{*{5}{p{0.09\textwidth}@{}}}
         $S = 0$ & 0.2 & 0.4 & 0.6 & 0.8
    \end{tabular} \\
    \rotatebox{90}{\scriptsize input}
    & \includegraphics[width=0.47\textwidth]{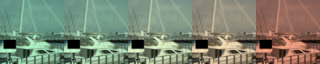} \\
    \rotatebox{90}{\scriptsize original}
    & \includegraphics[width=0.47\textwidth]{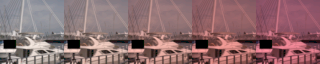} \\
    \rotatebox{90}{\scriptsize equivariant}
    & \includegraphics[width=0.47\textwidth]{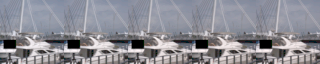} \\
    \end{tabular}
    \caption{Comparison of color corrections obtained by removing the illuminant estimated by the original and the equivariant Cerberus. The image is taken from the NUS dataset and has been distorted by the introduction of an artificial illuminant with saturation varying from 0.0 (no distortion) to 0.8.  For rendering purposes, the values of the pixels have been stretched, and the sRGB gamma has been applied to all images.}
    \label{fig:nus-example}
\end{figure}

To assess the robustness of the trained models, we repeated the test by distorting the test images with the application of a global artificial illuminant of random hue, and variable saturation.  The procedure is the same we described for image recognition, with the difference that this time the images are in the device RGB space and therefore there is no need to remove the gamma.  Figure~\ref{fig:nus} shows the results obtained.  The accuracy of the equivariant version degrades very slowly and remains quite good even for extreme illuminants (the median angular error is about 3.4 when the saturation is 0.9).
The original version, instead, clearly suffers the degradation of the input (at saturation 0.9 the median error is about 30 degrees). 
The stability of the equivariant color Cerberus is clearly visible in Figure~\ref{fig:nus-example}, where the output of the two variants at increasing levels of distortion are visually compared.

\subsection{Inpainting}
\label{sec:inpainting}

Image inpainting consists in filling the missing parts of an incomplete image, by introducing pixels that make the whole picture natural, and with a plausible content. The state of the art in image inpainting makes use of Generative Adversarial Networks (GAN).  Here we use the Context Encoder architecture proposed by Pathak \emph{et al.}~\cite{pathak2016context}.  Both the generator and the discriminator are sequential convolutional networks made of convolutions, ReLU, Leaky ReLU~\cite{maas2013rectifier} and batch normalization layers.  The generator ends with a final hyperbolic tangent, and the discriminator with a sigmoid.

We made offset equivariant the generator by replacing its components as described in Section~\ref{sec:equivariance}, and by adjusting the number of channels to the closest multiple of three. The input image is first converted in the logarithmic RGB space, and the final inpainted image is converted back in the linear RGB space.
Since the output values are already in the $[0, 1]$ range, they can be simply scaled to the range $[-1, 1]$ without the need for the final hyperbolic tangent.  For the discriminator we kept the original one without modifications.

\subsubsection{Paris Street View dataset}
\label{sec:paris}
In order to analyze the image inpainting task we considered the Paris Street View Dataset~\cite{doersch2012makes}. The images contained in the dataset have been extracted by the authors from Google Street View: a database composed of millions of street view images. Doersch \emph{et al.} extracted nearly \num{10000} images ($936 \times 537$) for each of the 12 cities considered (Paris, London, Prague, Barcelona, Milan, New York, Boston, Philadelphia, San Francisco, San Paulo, Mexico City, and Tokyo) and from suburbs of Paris.  Here we consider only the subset corresponding to the \num{6492} images of the city of Paris, divided in \num{6392} training and \num{100} test images.  As done in the original paper, all images have been resampled to $128 \times 128$ pixels, and the central $64 \times 64$ pixels have been erased.

We trained the original and the modified Context Encoder and we measured the performance on the test set.  Results are reported in Table~\ref{tab:paris} in terms of reconstruction error measured with the Peak Signal-to-Noise Ratio (PSNR). The performance of the two variants are very close and, for the case without distorted test images, they match those reported in the original paper.
\begin{table}
    \centering    
    \begin{tabular}{lcccc}
        \toprule
        & & \multicolumn{3}{c}{PSNR (dB)} \\
        Network & \# par. & $S~0.0$ & $S~0.5$ & $S~0.9$ \\
        \midrule
        Original & 71.14M & 17.65 & 18.44 & 19.05 \\  
        Equivariant & 69.94M & \textbf{17.67} & \textbf{18.68} & \textbf{19.78} \\  
        \bottomrule
    \end{tabular}
    \caption{Comparison of original and offset equivariant Context Encoder for image inpainting.
    Performance are measured in terms of PSNR (higher is better).
    Three levels of distortions are considered: no distortion ($S=0.0$), moderate  ($S=0.5$) and strong distortion ($S=0.9$).
    }
    \label{tab:paris}
\end{table}

\begin{figure}
    \centering
    \includegraphics[width=\linewidth]{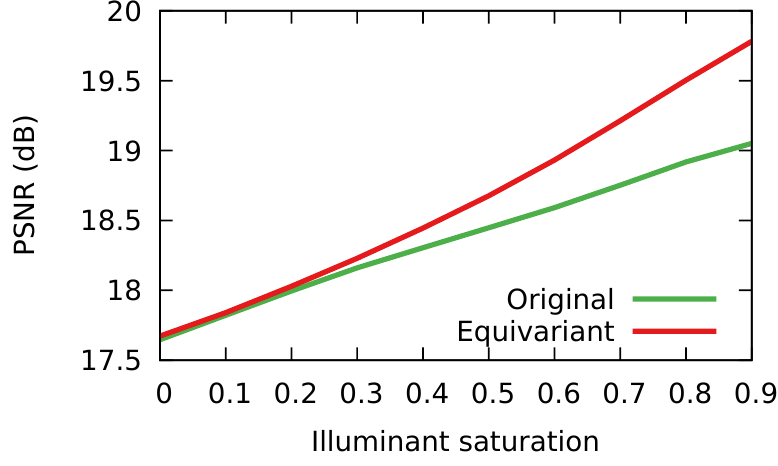}
    \caption{Reconstruction error (PSNR) obtained on the test images of the Paris Street View dataset, under an illuminant with random hue and variable saturation.}
    \label{fig:paris}
\end{figure}

We repeated the test by distorting the test images with the application of a global artificial illuminant of random hue, and variable saturation.  The procedure is the same used for image recognition and illuminant estimation.  Figure~\ref{fig:paris} shows the results obtained.  In this case performance improve with the level of saturation, but this is just an effect of the reduction in dynamic range caused by highly saturated illuminants.  Nevertheless, the offset equivariant version outperforms the original version for high values of saturation. 

\begin{figure}
    \centering
    \begin{tabular}{@{}r@{\hspace{1pt}}l@{}}
    & \begin{tabular}{*{5}{p{0.09\textwidth}@{}}}
         $S = 0$ & 0.2 & 0.4 & 0.6 & 0.8
    \end{tabular} \\
    \rotatebox{90}{\scriptsize real}
    & \includegraphics[width=0.47\textwidth]{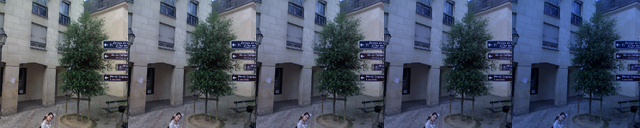} \\
    \rotatebox{90}{\scriptsize cropped}
    & \includegraphics[width=0.47\textwidth]{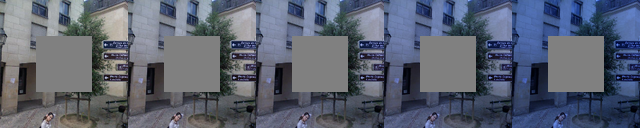} \\
    \rotatebox{90}{\scriptsize original}
    & \includegraphics[width=0.47\textwidth]{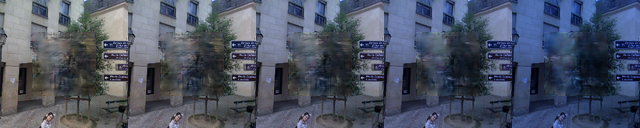} \\
    \rotatebox{90}{\scriptsize equivariant}
    & \includegraphics[width=0.47\textwidth]{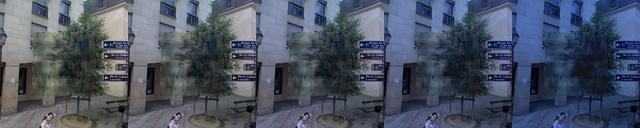} \\
    \end{tabular}
    \caption{Comparison of image inpainting obtained with the original and the equivariant Context Encoder. The image is taken from the Paris Street View dataset and has been distorted by the introduction of an artificial illuminant with saturation varying from 0.0 (no distortion) to 0.8.}
    \label{fig:paris-example}
\end{figure}

Figure~\ref{fig:paris-example} shows the behavior of the two models on one of the test images.  The nature of the problem makes it so both variants tend to follow the color distribution induced by the illuminant.  However, the equivariant version fills the center exactly with the same pattern, while the original version outputs different patterns as the saturation changes.

\section{Conclusions}
\label{sec:conclusions}
In this paper we defined the concept of offset equivariant networks. We proposed a framework to build them and showed how they can be used to achieve invariance or equivariance with respect to changes in the color of the illuminant in a variety of scenarios.  In all the experiments offset equivariant networks preserved the performance of the original architectures in standard conditions, and improved them under strongly non-neutral illuminants.
An implementation of the framework is publicly available at the  
address     \url{https://github.com/claudio-unipv/offset-equivariant}.

The consistency of offset equivariant networks is not learned from data, but it is guaranteed by their mathematical structure.  For this reason they represents a preferable alternative to other approaches (e.g. photometric data augmentation) in all those cases in which robustness to rare and unusual lighting conditions is a key requirement.

In this paper we focused on color images and on equivariance with respect to changes in the color of the illuminant.  In the future we plan to investigate if and how this kind of networks can be applied to other domains, such as multi-spectral and hyper-spectral imaging.


\bibliographystyle{elsarticle-num}
\bibliography{bibliography}

\end{document}